\documentclass[sigconf]{acmart}

\usepackage{balance}
\usepackage{multirow}
\usepackage{amsmath}
\usepackage{amssymb}
\usepackage{booktabs}

\def\BibTeX{{\rm B\kern-.05em{\sc i\kern-.025em b}\kern-.08emT\kern-.1667em\lower.7ex\hbox{E}\kern-.125emX}}
    
\copyrightyear{2019}
\acmYear{2019}
\setcopyright{acmlicensed}
\acmConference[MM '19] {Proceedings of the 27th ACM International Conference on Multimedia}{October 21--25, 2019}{Nice, France}
\acmBooktitle{Proceedings of the 27th ACM International Conference on Multimedia (MM'19), Oct. 21--25, 2019, Nice, France}
\acmPrice{15.00}
\acmDOI{10.1145/3343031.3351000}
\acmISBN{978-1-4503-6889-6/19/10}

\begin{document}

\fancyhead{}

\title{TGG: Transferable Graph Generation for Zero-shot and Few-shot Learning}


\author{Chenrui Zhang*\textsuperscript{1}, Xiaoqing Lyu*\textsuperscript{1} and Zhi Tang\textsuperscript{1,2}}
\affiliation{%
	\institution{\textsuperscript{1}Institute of Computer Science and Technology, Peking University, Beijing, China}
}
\affiliation{%
	\institution{\textsuperscript{2}State Key Laboratory of Digital Publishing Technology, Beijing, China}
}
\email{*{chenrui.zhang, lvxiaoqing}@pku.edu.cn}

\begin{abstract}
Zero-shot and few-shot learning aim to improve generalization to unseen concepts, which are promising in many realistic scenarios. Due to the lack of data in unseen domain, relation modeling between seen and unseen domains is vital for knowledge transfer in these tasks. Most existing methods capture seen-unseen relation \textit{implicitly} via semantic embedding or feature generation, resulting in inadequate use of relation and some issues remain (e.g. domain shift). To tackle these challenges, we propose a \textbf{Transferable Graph Generation} (\textbf{TGG}) approach, in which the relation is modeled and utilized \textit{explicitly} via graph generation. Specifically, our proposed TGG contains two main components: (1) \textbf{Graph generation} for relation modeling. An \textit{attention-based aggregate network} and a \textit{relation kernel} are proposed, which generate instance-level graph based on a class-level prototype graph and visual features. Proximity information aggregating is guided by a multi-head graph attention mechanism, where seen and unseen features synthesized by GAN are revised as node embeddings. The relation kernel further generates edges with GCN and graph kernel method, to capture instance-level topological structure while tackling data imbalance and noise. (2) \textbf{Relation propagation} for relation utilization. A \textit{dual relation propagation} approach is proposed, where relations captured by the generated graph are separately propagated from the seen and unseen subgraphs. The two propagations learn from each other in a dual learning fashion, which performs as an adaptation way for mitigating domain shift. All components are jointly optimized with a meta-learning strategy, and our TGG acts as an end-to-end framework unifying conventional zero-shot, generalized zero-shot and few-shot learning. Extensive experiments demonstrate that it consistently surpasses existing methods of the above three fields by a significant margin.
\end{abstract}


%

%

%
\maketitle

\vspace{-0.3em}
\section{Introduction}
In the past decade, traditional supervised learning has advanced rapidly due to deep learning techniques and large-scale labeled datasets. However, towards an ultimate machine learning paradigm, supervised learning is far from satisfactory in various real-world situations. On the one hand, the heavy reliance on large-scale labeled data makes it unscalable, as annotating sufficient data is laborious and costly, as well as the instances in some classes are quite rare for a long-tailed data distribution. On the other hand, supervised learning cannot deal with recognition tasks with ever-growing novel classes, which is urgently needed in many realistic scenarios.

To tackle the challenges stated above, Zero-Shot Learning (ZSL) and Few-Shot Learning (FSL) have recently emerged~\cite{frome2013devise,xian2017zero,vinyals2016matching,Sch2018Generalized}. Typically, ZSL aims to recognize unseen classes with no labeled instances during training, while a few representative instances of unseen classes are provided in FSL.
The key to the success of ZSL/FSL is the relation modeling between seen and unseen domains, which transfers knowledge from the seen domain to the unseen domain, for improving model's generalization to novel concepts.

Previous ZSL methods mainly focus on semantic embedding~\cite{akata2013label,zhang2015zero}, which learn a projection between visual space and semantic space. The principle of this paradigm is to utilize the side information (e.g., attributes or word vectors) shared by seen and unseen domains for projection learning, and measure similarity in the resulting semantic space for final classification. Such a projection-based paradigm is limited by the \textit{heterogeneity} between visual feature and side information, as well as the \textit{domain shift}~\cite{romera2015embarrassingly} when the learned projection is directly applied to unseen domain without adaptation. Moreover, each class is represented as a fixed embedding point in semantic space, while the intra-class variation and discriminative information implied in visual data distribution are ignored~\cite{wang2018zero}.

Recently, deep generative models have been introduced as alternative frameworks in ZSL~\cite{zhang2018visual,huang2018generative,fvaegan}. In this paradigm, visual feature and side information of seen domain are utilized for capturing visual-semantic joint distribution, and then the visual feature of unseen domain can be synthesized conditioned on the associated side information. Hence, ZSL can be converted into a supervised problem, as the synthesized visual features can be straightforwardly fed to typical classifiers for supervised training. However, the inherent handicap of this paradigm is that evaluating how well the dummy features capture the targeted unseen domain distribution is still ambiguous. Furthermore, the instability of generative models (e.g., mode collapse of generative adversarial networks~\cite{Goodfellow2014Generative}) leads to noisy synthesized feature with poor diversity, which is harmful for the downstream classifier training.

Paradigms stated above fall under the taxonomy of implicit relation modeling methods, in which the use of relation is inadequate and some key issues (e.g. domain shift) are still unsolved. In contrast, another novel paradigm is proposed~\cite{wangxiaolong2018zero} to explicitly utilize knowledge from knowledge graph (KG) for ZSL. Typically, these methods are built upon the graph convolutional network (GCN)~\cite{Kipf2016Semi}, which distills knowledge from KG for class-level relation modeling. The graph nodes denote the class embedding, while the edges describe the relations of different classes. Despite the promising performance, they still have some shortcomings. First, they simply learn an independent classifier for each class, while the unseen class labels are not involved and thus domain shift remains. Second, the relation is only modeled at class level, while the instance-level relation is ignored, resulting in the loss of discriminative ability. Third, the utilization of relation in these methods is still implicit, where the distilled knowledge can get diluted during classification.

To overcome the above limitations, in this paper, we propose to explicitly model and utilize relation at both class level and instance level, via graph generation and relation propagation. Specifically, we propose a \textbf{Transferable Graph Generation} (\textbf{TGG}) approach, which contains a graph generation module and a relation propagation module. The details are presented as follows.

\textbf{Graph generation module} aims to capture relations among class concepts, attributes and visual instances. In this module, an attention-based aggregate network and a relation kernel are proposed, which take a class-level prototype graph and visual instances as inputs, and output instance-level graphs with the revised instance embeddings as nodes and their relations as edges. The prototype graph is derived from an off-the-shelf knowledge graph, which acts as a relation template and will be enriched by integrating visual information during graph generation. In order to model comprehensive seen-unseen relation and reduce domain gap, we introduce unseen information at both class level and instance level from the very beginning. For class level, the prototype graph is constructed to contain class concepts of both seen and unseen domains. For instance level, the graph generation module is also fed with instances of both domains, here we skillfully unify the ZSL and FSL with the dummy feature synthesis. Concretely, we synthesize dummy features for unseen classes via Generative Adversarial Networks (GANs)~\cite{Goodfellow2014Generative}, and they will be treated equally as the few provided instances in FSL. Hence, the graph generation can be fully-supervised, which is beneficial for the downstream relation utilization. 

Our aggregate network aims to learn a revised node embedding space, which revises the input visual features by aggregating neighbors' information at both class and instance level. A multi-head graph attention mechanism is proposed to enhance the aggregation procedure, which prevents information dilution and negative knowledge transfer. The relation kernel is proposed to explicitly generate relations/edges over the revised nodes, where GCNs and graph kernel methods are used to tackle data imbalance and noise.

\textbf{Relation propagation module} aims to make full use of the learned relations for final classification. Compared with the implicit embedding methods, knowledge transfer in graph manifold space with explicit relation inference is more efficient, and it helps to learn better decision boundary. Motivated by this, and with the advantages of fully-supervised graph generation, we propose a \textit{dual relation propagation} approach, to explicitly infer supervision via relation propagation and further alleviate domain shift with dual learning. Relations in the generated graph start propagation separately from seen and unseen subgraphs, and the two reverse propagations learn from each other in a dual learning manner.

Moreover, we joint optimize all the above components end-to-end with an episodic training strategy of meta-learning. Graph nodes of both seen and unseen classes are randomly divided into training and test subsets, where relations are used for missing label prediction. Such a strategy ensures that the settings of training and test in our TGG are consistent, reducing inductive bias significantly.

The main contributions are summarized as follows:
\vspace{-0.5em}
\begin{itemize}
	\item We propose a Transferable Graph Generation (TGG) approach, to explicitly model and utilize seen-unseen relation for ZSL/FSL via graph generation. We design an attention-based aggregate network and a relation kernel, which capture multi-granular relations and are robust to data imbalance and dummy data noise.
	\item We propose a dual relation propagation approach to utilize relation explicitly, which alleviates domain shift with fully-supervised relation propagation in a dual learning manner. An episodic training strategy is designed based on meta-learning, combining all components of our TGG for end-to-end joint optimization.
	\item Our TGG acts as a unified framework for conventional zero-shot, generalized zero-shot and few-shot learning, and as demonstrated by extensive experiments, it consistently outperforms existing methods by a large margin. The code of our work is available at: \url{https://github.com/zcrwind/tgg-pytorch}.
\end{itemize}

\section{Related work}

\begin{figure*}[htb]
	\begin{center}
		\includegraphics[height=4.691cm]{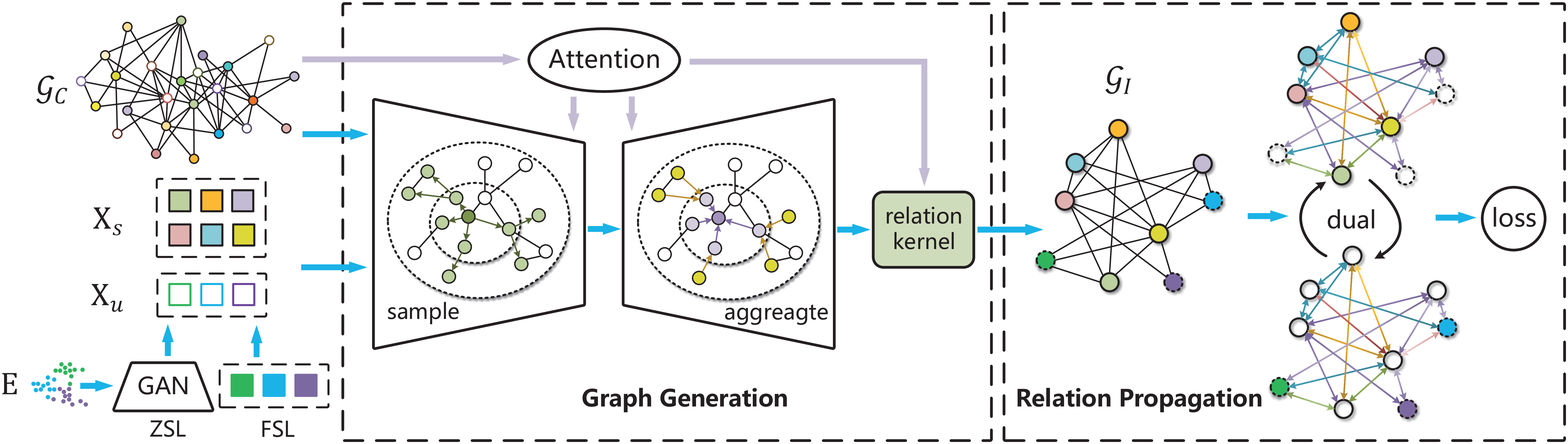}
	\end{center}
	\vspace{-1.047em}
	\caption{Architecture of the proposed TGG. $\mathcal{G}_C$ and $\mathcal{G}_I$ denote the class-level and instance-level graph, respectively. $X_s$ and $X_u$ is the instances of seen and unseen classes, respectively. $E$ is the side information.}
	\label{fig:framework}
	\vspace{-0.89em}
\end{figure*}

\subsection{Zero-shot and few-shot learning}

\subsubsection{Zero-shot learning (ZSL)}
According to the label space setting for evaluation, existing ZSL methods can be divided into two categories, i.e. conventional ZSL and generalized ZSL (GZSL)~\cite{xian2017zero}. Conventional ZSL aims to learn classifier based on seen instances, and then evaluate the trained model on unseen instances, where the label spaces of seen and unseen domains are totally disjoint, and model evaluation is only performed on the unseen domain. In contrast, GZSL aims to classify instances in the combination of seen and unseen classes, which is more realistic in practice.

From the algorithm perspective, existing ZSL and GZSL methods can be grouped into three paradigms~\cite{wangxiaolong2018zero,huang2018generative}. The first paradigm, known as semantic embedding, learns a projection between visual and semantic space with the aid of side information, then the learned projection will be applied directly to the unseen domain during test, where unseen instances can be classified by certain similarity measurement~\cite{akata2013label,zhang2015zero}. Due to the heterogeneity between visual and semantic feature, such a paradigm suffers from the information degradation issue. Recently, attention has shifted to another paradigm, which uses generative models to synthesize unseen feature. Huang et al. \cite{huang2018generative} utilize GANs~\cite{Goodfellow2014Generative} to learn visual-semantic joint distribution, and unseen instances can be synthesized as dummy data, which are used to convert ZSL to a typical supervised problem. In contrast to the above paradigms, a new paradigm is rising lately for borrowing power from structure knowledge. \cite{wangxiaolong2018zero} and \cite{Kampffmeyer2018Rethinking} use knowledge graph and GCNs to predict classifiers for each class. The constraint is a mean-square error between the predicted and ground truth classifiers of seen classes. While promising, there are mainly two shortcomings of them. First, the generalization is limited by the fixed ground truth classifiers. Second, the relation only focuses on the seen domain at class-level. Our TGG captures relation among seen and unseen classes at both class-level and instance-level.

\subsubsection{Few-shot learning (FSL)}
The data sparsity issue leads the typical finetuning strategy not adaptable for FSL, as overfitting is easy to happen. Thus, current FSL research turns to meta-learning, a new supervised learning setup that performs optimization over batches of tasks rather batches of data. The task that meta-learner tries to solve, called \textit{episode task}, corresponds to independent learning problem that simulates the few-shot setting within episodes, and thus helps to learn high generalization. 
Siamese Networks~\cite{koch2015siamese} learn pair-wise distance under the principle that similar instances should be close, and then perform one-shot classification by nearest neighbors search. Matching network~\cite{vinyals2016matching} is an end-to-end trainable k-nearest neighbors framework for FSL, in which the pair-wise distance is computed by cosine similarity. Prototypical network~\cite{snell2017prototypical} extends~\cite{vinyals2016matching} by replacing cosine distance with Euclidean distance, and learns class prototypes for similarity measurement.
However, meta-learning based models typically cannot scale to ZSL, and we address this limitation via graph generation.

\subsection{Graph learning}
Our work is conceptually related to graph neural networks (GNNs) w.r.t. architecture, as well as graph generation w.r.t. application.

GNNs are first introduced by \cite{Gori2005A,Franco2009The}, whose target is to learn a state embedding that contains neighborhood information for each node in graphs. In \cite{Gori2005A,Franco2009The}, a parametric local transition function is applied on all nodes in a stacked manner, where a recurrent message propagation is learned discriminatively. \cite{li2015gated} proposes the gated graph neural network (GGNN), which uses the Gate Recurrent Units (GRU) in the propagation step to untie the recurrent layer weights, and increase nonlinearity via gate mechanism. Bruna et al. \cite{bruna2013spectral} propose to learn spectral convolution in the Fourier domain via eigen decomposition on the graph Laplacian. Subsequent work \cite{defferrard2016convolutional} reduces the computational complexity of \cite{bruna2013spectral} by learning polynomials of the graph Laplacian.
As one of the most representative graph convolutional networks, GCN~\cite{Kipf2016Semi} is proposed to solve the semi-supervised problem via spectral methods, which learns layer-wise propagation operations directly on graphs. GraphSAGE~\cite{Hamilton2017Inductive} acts as a spatial graph convolutional method, which uniformly samples a fixed number of neighbors for each node, and then uses different aggregating functions for large graph node embedding.

Recently there has been a surge of interest in graph generation, due to its wide applications on molecule discovery, social network analysis and knowledge graph construction. NetGAN~\cite{Bojchevski2018NetGAN} has done a preliminary trial on graph generation via random walk, which converts graph generation to a walk sequence generation problem via generative adversarial training~\cite{Goodfellow2014Generative}. MolGAN~\cite{Cao2018MolGAN} utilizes GAN~\cite{Goodfellow2014Generative} and reinforcement learning (RL) to generate discrete graph structure, where a permutation-invariant discriminator is designed to handle the node variant, as well as a RL-based reward function is developed to endow the generated molecule with the desired chemical properties. Li et al.~\cite{li2018learning} propose to generate graph nodes and edges sequentially, where GNNs are applied to learn latent states of current graph, and then the latent states will be used as the history memory for deciding the next generation action.

All the above graph generation methods are based on the fact that there exists real graph data for distribution fitting, while our work focuses on generating graph without prior distribution information, and can be generalized to unseen node types.

\section{Our TGG Approach}
As illustrated in Figure~\ref{fig:framework}, our Transferable Graph Generation (TGG) framework mainly contains two components, i.e. \textit{graph generation} and \textit{relation propagation}. Graph generation module takes the class-level graph and real/dummy visual instances of seen/unseen classes as input, learns both node embeddings and relations with aggregate network and relation kernel. Relation propagation module exploits generated relation graph for classification, via a dual relation propagation approach with meta-learning strategy.

\subsection{Preliminaries}
\begin{figure}[htb]
	\begin{center}
		\includegraphics[height=3.06cm]{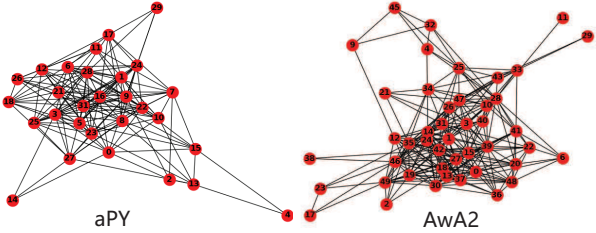}
	\end{center}
	\vspace{-1.1em}
	\caption{Class-level graphs of aPY and AwA2 dataset.}
	\label{fig:class_level_graph}
	\vspace{-1.3em}
\end{figure}
\subsubsection{Problem Formulation}
Let $D_{tr}=\{\left(x_i,e_i,y_i\right)\}_{i=1}^{N_s}$ denote the training set of $ N_s $ image instances, and $ D_{te}=\{\left(x_i,e_i,y_i\right)\}_{i=1}^{N_u} $ denote the test set of $ N_u $ image instances. Their corresponding label spaces are $ {Y^S}=\{1,2,\cdots,S\} $ and $ {Y^U}=\{S+1,S+2,\cdots,S+U\} $ with ${Y^S}\cap{Y^U}=\varnothing$. $S$ and
$U$ here denote the total number of seen and unseen classes, respectively. $x_i\in\mathbb R^{d}$ is the $d$-dimensional visual feature of the $i$-th instance with label $y_i$, and $e_i\in\mathbb R^m$ denotes the side information (e.g., attributes or word vectors) uniquely associated with the class label $y_i$. Based on the symbol definition, we then formulate three problems addressed in this paper as below.
\begin{itemize}
	\item \textit{Zero-shot Learning (ZSL)}: The image features of unseen classes $Y^U$ are not available during training. The goal of ZSL is to predict the label $y_u\in{Y^U}$ given an unseen class instance using its visual feature $x_u$.
	\item \textit{Generalized Zero-shot Learning (GZSL)}: The image features of unseen classes $Y^U$ are not available during training. The goal of GZSL is to predict the label $l\in{Y^S}\cup{Y^U}$ given an image instance using its visual feature $x$.
	\item \textit{Few-shot Learning (FSL)}: Only a few/one randomly chosen image instances from unseen classes $Y^U$ are available with label information during training, and the goal of FSL is same with ZSL and GZSL settings above.
\end{itemize}

\subsubsection{Class-level graph construction}\label{sec:class_level_graph_construction}
Similar to~\cite{gao2019know}, we exploit ConceptNet 5.5~\cite{speer2017conceptnet} for class-level graph construction, which is an off-the-shelf knowledge graph connecting words and phrases of natural language edges. It is noted that we treat CUB dataset~\cite{wah2011caltech} (see Section~\ref{sec:experiments}) as a special case, as its class labels are proper nouns of fine-grained birds, which is hard to build semantic connections via ConceptNet. Rather we build the class-level graph of CUB via computing Hadamard product over part-level attributes. The resulting graph is densely connected with normalized edge weights, which denote similarities among different classes. The class-level graphs of two small datasets are shown as examples in Figure~\ref{fig:class_level_graph}.

\subsubsection{Dummy visual feature synthesis}
For ZSL and GZSL, we synthesize dummy visual feature for unseen classes, using the recently emerged generative adversarial learning~\cite{Goodfellow2014Generative}. Specifically, we use conditional GAN~\cite{mirza2014conditional} to perform $semantic \rightarrow visual$ synthesis conditioned on the associated side information, and use WGAN-GP~\cite{gulrajani2017improved} for training settings. To stabilize the training of GAN, similar to~\cite{zhang2018visual,huang2018generative}, a dual learning mechanism is applied with semantic feature regression. We use visual feature synthesis as a pre-processing step rather directly learning relations over it, as we believe that there are several unsolved issues with such feature synthesis methods for ZSL/GZSL. First, the generated feature cannot fit the true distribution very well, and is thus suboptimal for GZSL. Second, instance-level relations cannot be captured by the generated feature in such feature mapping learning, where intra-class variance is ignored.
Our TGG revises them into a node embedding space via explicit relation modeling.

\subsection{Graph Generation}
\label{sec:graph_generation}
\subsubsection{Attention-based aggregate network}

As shown in Figure~\ref{fig:framework}, the graph generation module of our TGG takes the class-level graph and visual feature as inputs, where the synthesized dummy feature is used for unseen classes in ZSL/GZSL, and the few provided unseen class features are used repeatedly in FSL. Our goal of graph generation is to generate implicit instance representations as node embeddings and explicit relations as edges, via incorporating proximity information from each node's neighborhood at both class level and instance level.

We draw inspiration from GraphSAGE~\cite{Hamilton2017Inductive}, an inductive variant of GCN~\cite{Kipf2016Semi}, to develop our aggregate network. The core operations of GraphSAGE can be formulated as follows:
\begin{equation}
h^k_{\mathcal N(v)}\leftarrow\text{AGGREGATE}_k\big(\{h_u^{k-1},\forall u\in \mathcal N(v)\}\big)
\label{eq:graphsage_1}
\end{equation}
\begin{equation}
h^k_{v}\leftarrow \sigma\big(\textbf{W}^k\cdot\text{CONCAT}(h_v^{k-1},h_{\mathcal N(v)}^{k-1})\big)
\label{eq:graphsage_2}
\end{equation}
where $\text{AGGREGATE}_k$ denotes the aggregation function at k-hop, which aggregates neighbor information for the subsequent node embedding update. $v$ and $u$ are nodes in the graph $\mathcal{G}(\mathcal{V},\mathcal{E})$, here $\mathcal{V}$ and $\mathcal{E}$ denote node and edge set of $\mathcal{G}$, respectively. $h_v^k$ is the node embedding of source node $v$ at k-th propagation, and $\mathcal N$ denotes neighbor sampling function: $\mathcal{N}(v):v\rightarrow 2^\mathcal{V}$. After information aggregation, node embedding of $v$ and its neighbors $h^k_{\mathcal N(v)}$ will be concatenated via $\text{CONCAT}$ operation and activated by $\sigma$ non-linearity, in which the trainable weights $\textbf{W}^k$ can be learned.

As shown in Eq.(\ref{eq:graphsage_1}) and Eq.(\ref{eq:graphsage_2}), neighbor sampling and aggregation are two main components in GraphSAGE. In terms of sampling, GraphSAGE uniformly samples neighbors with fixed numbers. As for aggregation, GraphSAGE explores three kinds of aggregation functions, namely mean, LSTM and pooling. Mean aggregation simply averages over all neighbor node features, while LSTM and pooling workarounds integrate node features via LSTM architecture or pooling operation. However, we argue that these mechanisms are insufficient in our graph generation situation for ZSL/FSL, as the generated graph should integrate proximity information more precisely, to cope with noise of dummy features and prevent negative knowledge transfer. Furthermore, our TGG performs graph learning over graphs of different granularity, namely class-level prototype graph $\mathcal{G}_C$ and instance-level graph $\mathcal{G}_I$, thus, uniform operations might loss discriminative information in such $\mathcal{G}_C\rightarrow\mathcal{G}_I$ graph translation procedure.

To solve the above issues, we propose to enhance GraphSAGE algorithm with a multi-head attention mechanism~\cite{Veli2017Graph}. Concretely, we design class-level and instance-level attention during aggregation, and combine them analogous to multiple channels in ConvNet. The instance-level attention is defined as follows:

\begin{equation}
z_i^{k-1}=\textbf{W}^{k-1}h_i^{k-1}, \forall i\in\mathcal{V}
\end{equation}
\begin{equation}
e_{vu}^{k-1}=\text{LeakyReLU}\big(\vec {A}^{k-1^T}\!\!\!\cdot\text{CONCAT}(z_v^{k-1},z_u^{k-1})\big)
\label{eq:unnormalized_att}
\end{equation}
\begin{equation}
\alpha_{vu}^{k-1}=\frac{\exp(e_{vu}^{k-1})}{\sum_{j\in \mathcal{N}(v)}^{}\exp(e_{vj}^{k-1})}
\label{eq:softmax_att}
\end{equation}
\begin{equation}
h_v^{k}=\sigma\Big(\!\sum_{j\in \mathcal{N}(v)} {\alpha^{k-1}_{vj} z^{k-1}_j }\Big)
\label{eq:insatnce_att_final}
\end{equation}
where $z_i^{k-1}$ is first obtained by performing linear transformation on the node embedding from the last aggregation $h_i^{k-1}$, then a pair-wise additive attention score between two neighbors is computed as $e_{vu}^{k-1}$ in Eq.(\ref{eq:unnormalized_att}), which concatenates $z_v^{k-1}$ and $z_u^{k-1}$ first, then takes dot product between the concatenation and a trainable weight vector $\vec{A}^{k-1^T}$, followed by a $\text{LeakyReLU}$ non-linearity. Next, Eq.(\ref{eq:softmax_att}) locally normalizes the attention scores over each node's neighbors. Finally, in Eq.(\ref{eq:insatnce_att_final}), aggregation similar to Eq.(\ref{eq:graphsage_2}) is performed over neighbor embeddings according to the attention score.

In another vein, the class-level attention score can be derived directly from the weights of $\mathcal{G}_C$ (see Section~\ref{sec:class_level_graph_construction}), and we just normalize them in each local aggregation like Eq.(\ref{eq:softmax_att}). Instance-level and class-level attention have independent parameters and we combine them as a multi-head attention form by:
\begin{equation}
h_v^{k}=\sigma\Big(\frac{1}{|a|}\sum_{a}\!\sum_{j\in \mathcal{N}(v)} {\alpha^{k-1}_{vj} \textbf{W}^{k-1} h^{k-1}_j }\Big)
\label{eq:multi_head_att}
\end{equation}
here $a\in\{att_C,att_I\}$ denotes the attention type with candidates of class-level attention $att_C$ and instance-level attention $att_I$. The motivation behind Eq.(\ref{eq:multi_head_att}) is that weighting neighbor features with multi-level attention helps aggregate proximity information more precisely and efficiently, which deals with information dilution~\cite{Kampffmeyer2018Rethinking}, and is vital for ZSL/FSL generalization when faced with a lack or noise of data. Moreover, such node embedding revision is performed among the seen and unseen classes, in which the distribution of two domains tends to be consistent via neighbor information integration, and thus domain gap can be reduced significantly to alleviate the domain shift issue.

\subsubsection{Relation kernel}
\begin{figure}[htb]
	\begin{center}
		\includegraphics[height=3.72cm]{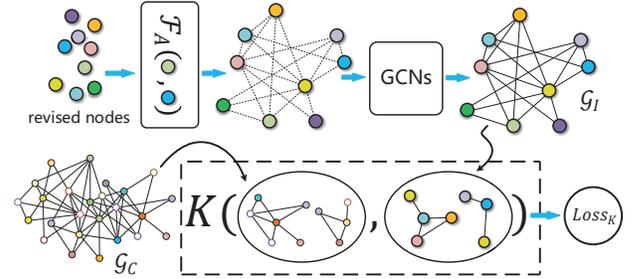}
	\end{center}
	\vspace{-0.8em}
	\caption{Relation kernel of our TGG. $\mathcal{F}_A$ is the edge feature learning function, $K$ denotes the graph kernel.}
	\label{fig:relation_kernel}
	\vspace{-1.0em}
\end{figure}

The aggregate network above revises node embeddings over both seen and unseen classes, by integrating proximity knowledge in an implicit manner. Based on this revised node embedding space, we further generate relations explicitly, to exploit graph manifold for better seen-to-unseen generalization. To this end, we propose a relation kernel module (Figure~\ref{fig:relation_kernel}), to explicitly learn edge features and thus generate instance-level graphs.

Taking the permutation invariance and distance properties (e.g., identity) into account, we first design edge feature learning function as:
\begin{equation}
A^{k}_{vu}=\text{exp}\Big(-\frac{\Phi_{\Theta}\big(\text{abs}\,(h_v^{k}-h_u^{k})\big)}{2\delta^2}\Big)
\label{eq:edge_learning}
\end{equation}
where $A^{k}_{vu}$ denotes the generated edge between node $v$ and $u$ in the adjacency matrix $A$ of $\mathcal{G}_I$, $\Phi_{\Theta}$ is a neural network parameterized with $\Theta$, and $\delta$ is a bandwidth hyperparameter. Mathematically, Eq.(\ref{eq:edge_learning}) is an instantiation of Gaussian similarity function with Manhattan distance, yielding learnable edge features with $\Phi_{\Theta}$. Once $A$ is obtained, it will be fed into stacked GCN modules for graph generation:
\begin{equation}
H^{(l)}=\sigma\big(\tilde{D}^{-\frac{1}{2}}\tilde{A}\tilde{D}^{-\frac{1}{2}}H^{(l-1)}\textbf{W}^{(l)}\big)
\end{equation}
here $\tilde{A}=A+I$ is obtained by add self-connections on $A$, $I$ is the identity matrix, $\tilde{D}_{ii}=\sum_j\tilde{A}_{ij}$, and $\textbf{W}^{(l)}$ is the trainable filter in the $l$-th layer of GCN.

Furthermore, an additional graph regularization item is designed in our relation kernel, which is optimized jointly with the downstream classification task:
\begin{equation}
Loss_K(\mathcal{G}_C,\mathcal{G}_I)=GraphKernel(A^{L},\mathcal{G}_C^*)
\label{eq:regular_item}
\end{equation}
here $A^{L}$ is the final learned $A$ in the $L$-th GCN layer, and $\mathcal{G}_C^*$ means the normalized subgraph of $\mathcal{G}_C$ that shares node set with $A^L$. $GraphKernel(\cdot)$ is the graph kernel that measures graph similarities via computing global graph representations. In this work, we use graph2vec~\cite{narayanan2017graph2vec} as the graph kernel, which is task agnostic and can be learned in an unsupervised manner. Eq.(\ref{eq:regular_item}) ensures the generated local relations in $\mathcal{G}_I$ are consistent with the similarities derived from $\mathcal{G}_C$, which aids zero-shot relation generation as a priori information and overcomes overfitting.

\subsection{Relation Propagation}
Once the instance-level graph $\mathcal{G}_I$ is generated (Section~\ref{sec:graph_generation}), the node embedding and relations can be utilized for ZSL/FSL classification. To make full use of the knowledge within $\mathcal{G}_I$, we propose to explicitly perform relation inference with a novel dual relation propagation and meta-learning, as presented below.
\subsubsection{Dual relation propagation}
To explicitly utilize the learned relations for improving generalization and further alleviating domain shift, we propose a \textbf{dual relation propagation} between the seen and unseen subgraphs in $\mathcal{G}_I$. Specifically, we evolve standard label propagation algorithm~\cite{zhu2002learning} by inter-domain dual learning. To keep this paper self-contained, we briefly review the standard label propagation algorithm, then elaborate our dual relation propagation between seen and unseen subgraphs.

Label propagation (LP) is a classic algorithm for semi-supervised learning. Suppose $\{(x_1,y_1)\dots(x_l,y_l)\}$ be the labeled data, $y\in\{1\dots C\}$, and $\{(x_{l+1},y_{l+1})\dots(x_{l+u},y_{l+u})\}$ the unlabeled data. Let $\mathcal{Y}$ denote the set of $(l+u)\times C$ matrix. LP defines a label matrix $Y\in\mathcal{Y}$ with $Y_{ij}=1$ if $x_i$ is a labeled instance with label $y_i=j$, otherwise $Y_{ij}=0$. The goal of LP is to propagate the labels through pre-computed edges, to determine the unknown labels of instances in $Y$. LP has been proven~\cite{zhu2002learning} to have closed-form solution as:
\begin{equation}
Y^{*}=(I-\mu{Y^L})^{-1}Y
\end{equation}
where $I$ is the identity matrix, $Y^L$ is the labeled sub-matrix of $Y$, and $\mu\in(0,1)$ is a hyperparameter that controls the amount of propagated information.

As we introduce unseen domain information in two-folds, namely unseen prototype in class-level graph and dummy feature inputs (for ZSL setting), all nodes in the generated graph $\mathcal{G}_I$ are actually labeled. Based on such supervised setting advantages provided by graph generation, we propose dual relation propagation between seen and unseen domains. More concretely, we separately use seen and unseen instances as labeled data for label propagation, and make sure that the resulting label matrices are consistent. The constraint of our dual relation propagation is defined as:
\begin{equation}
Loss_d=\big\Vert\,(I-\mu{Y^S})^{-1}Y-(I-\mu{Y^U})^{-1}Y\,\big\Vert_F^2
\label{eq:dual_relation}
\end{equation}
where $Y^S$ and $Y^U$ denote the labeled sub-matrices of seen and unseen instances, respectively. $\Vert\cdot\Vert_F$ means the Frobenius norm of a matrix. Label propagations starting from the seen and unseen subgraph in $\mathcal{G}_I$ can be regarded as two propagation learners with reverse learning direction, and minimizing Eq.(\ref{eq:dual_relation}) encourages them to learn from each other `how to propagate'.

\subsubsection{Meta-learning based training strategy}
We now present how our TGG framework unifies FSL, ZSL and GZSL with meta-learning, where graph generation, relation propagation and final classification are jointly optimized in an end-to-end manner. For FSL, there are three datasets, namely training, testing and support sets. The testing and support sets share the same label space (i.e., unseen space), which is disjoint with the seen space of training set. Suppose the support set has $K$ labeled instances for each $N$ unique classes, the FSL task is called $N$-way, $K$-shot. As for ZSL, we borrow the power of conditional GAN~\cite{mirza2014conditional} to build a dummy support set for unseen classes, with the side information as conditions.

In traditional meta-learning paradigm with episodic training, each episode simulates the few-shot setting with a subset of the training set. In this paper, we follow the episodic training of meta-learning, but extend its label space during graph generation. Specifically, we involve the unseen prototype in $\mathcal{G}_C$, as well as input unseen class instances from the dummy/real support set in ZSL/FSL, thus the graph generation learning can pick neighbor information from both seen and unseen classes. Furthermore, another associated difference lies in that we also extend label space to the union of seen and unseen domains in episodic task simulation, towards a fully-supervised meta-learning for performance improvement.
Thanks to the introduction of dummy unseen instances and the use of graph learning for revising them, ZSL, GZSL and FSL can be solved in TGG uniformly, where graph generation, relation propagation and classification can be jointly optimized end-to-end with episodic training. As a result, domain shift and classifier bias to the seen domain can be reduced significantly (as shown in Section~\ref{sec:experiments}).

In each episode, we obtain the final predictions by normalizing the propagation results to probabilistic values with softmax:
\begin{equation}
P(\tilde{y}_i=j\,|x_i)=\frac{\text{exp}(Y^*_{ij})}{\sum_{p=1}^{N}\text{exp}(Y^*_{ip})}
\end{equation}
where $\tilde{y}_i$ is the predicted label for the test instance $x_i$, and $N$ is the class number in an episode to be classified. Then, we use cross-entropy for the final classification:
\begin{equation}
Loss_c=\sum_{i=1}^{N\times K+T}\sum_{j=1}^{N}-\mathbb{I}(y_i=j)P(\tilde{y}_i=j\,|x_i)
\end{equation}
where $\mathbb{I}(\cdot)$ is the indicator function and $y_i$ is the ground truth label for instance $x_i$. $N\times K+T$ is the instances number in a $N$-way $K$-shot episode with $T$ test instances. Comprehensively, the objective of our TGG is summarized as follows:
\begin{equation}
J(\theta_{TGG})=Loss_c+\lambda_1 Loss_d+\lambda_2 Loss_{K}\,(\mathcal{G}_C,\mathcal{G}_I)
\label{eq:total_loss}
\end{equation}

Essentially, we utilize the generated relations to learn a metric in graph manifold. That is, TGG learns a graph manifold metric in a revised node embedding space, rather pre-defining a fixed metric (e.g., Euclidean) in a projection space.
The reasons for applying meta-learning are three-folds.
First, traditional graph architectures such as GCN and GraphSAGE are hard to end-to-end solve ZSL and GZSL simultaneously, as the class number must be pre-defined as the output dimension in the last output layer.
Second, meta-learning actually performs as an adaptation method, which moves testing adaptation to training stage via episodic task simulation.
Third, such meta-learning settings can be utilized to further alleviate the domain shift, since it ensures the test and the train environments are consistent in our TGG.

\section{Experiments}\label{sec:experiments}

\begin{table}
	\caption{Statistics of datasets.}
	\vspace{-1.0em}
	\label{tab:datasets}
	\begin{tabular}{ccccccc}
		\toprule
		\multirow{2}{*}{Dataset}		&\multirow{2}{*}{\#att}	&\multicolumn{2}{c}{Class number} & \multicolumn{3}{c}{Image number} \\
		\cmidrule{3-4} \cmidrule{5-7}
		&  & \#$Y^S$ & \#$Y^U$ & Total & \#$D_{tr}$ & \#$D_{te}^{U}$/\#$D_{te}^{S}$ \\
		\midrule
		\textbf{aPY} 	& 64				& 15+5						& 12				&	15339			& 5932					& 7924/1483 \\
		\textbf{AwA2} 	& 85				& 27+13						& 10				&	37332			& 23527					& 7913/5882 \\
		\textbf{CUB} 	& 312	 			& 100+50					& 50				&	11788			& 7057					& 2679/1764	\\
		\textbf{SUN} 	& 102				& 580+65					& 72				&	14340			& 10320					& 1440/2580 \\
		\bottomrule
	\end{tabular}
	\vspace{-1.0em}
\end{table}

\begin{table*}
	\caption{Accuracy (\%) results of ZSL and GZSL evaluated on four benchmark datasets.}
	\vspace{-0.8em}
	\label{tab:ZSL_GZSL_results}
	\begin{tabular}{r|cccc|cccc|cccc|cccc}
		\toprule
		\textbf{Dataset}	& \multicolumn{4}{c|}{\textbf{aPY}} & \multicolumn{4}{c|}{\textbf{AwA2}} & \multicolumn{4}{c|}{\textbf{CUB}} & \multicolumn{4}{c}{\textbf{SUN}} \\
		\midrule
		\textbf{Methods} &ZSL	& U & S 	& HM 	&ZSL	& U 	& S 	& HM 	&ZSL  & U 	  & S 	  & HM 	 &ZSL	& U 	& S	 	& HM \\
		\midrule
		SSE~\cite{zhang2015zero}&34.0	 & 0.2 	& 78.9 		& 0.4	&61.0	& 8.1 	& 82.6 	& 14.8	&43.9  & 8.5   & 46.9  & 14.4 &	51.5 & 2.1 	 & 36.4 	& 4.0 \\
		LATEM~\cite{xian2016latent}&35.2 & 0.1  & 73.0 		& 0.2	&55.8	& 11.5  & 77.3 	& 20.0	&49.3  & 15.2  & 57.3  & 24.0 &	55.3 & 14.7  & 28.8 	& 19.5 \\
		ALE~\cite{akata2013label}&39.7   & 4.6  & 73.7 		& 8.7	&62.5	& 14.0  & 81.8 	& 23.9	&54.9  & 27.3  & 62.8  & 34.4 &	58.1 & 21.8  & 33.1 	& 26.3 \\
		DEVISE~\cite{frome2013devise}&\textbf{39.8}	& 4.9  	& 76.9 	& 9.2	&59.7	& 17.1  & 74.7 	& 27.8	&52.0  & 23.8  & 53.0  & 32.8 & 56.5 & 16.9  & 27.4 	& 20.9 \\
		SJE~\cite{akata2015evaluation}&32.9	& 3.7  	& 55.7 	& 6.9	&61.9	& 8.0   & 73.9 	& 14.4	&53.9  & 23.5  & 52.9  & 33.6 &	53.7 & 14.7  & 30.5 	& 19.8 \\
		ESZSL~\cite{romera2015embarrassingly}&38.3	& 2.4  	& 70.1 	& 4.6	& 58.6	& 5.9   & 77.8 	& 11.0	& 53.9 & 12.6  & 63.8  & 21.0 &	54.5 & 11.0  & 27.9 	& 15.8 \\
		SYNC~\cite{changpinyo2016synthesized}&23.9	& 7.4 	& 66.3 	& 13.3	& 46.6	& 10.0  & 90.5 	& 18.0	& 55.6 & 11.5  & \textbf{70.9} & 19.8 &	56.3 & 7.9   & 43.3 	& 13.4 \\
		SAE~\cite{kodirov2017semantic}&34.0	& 0.4   & \textbf{80.9} & 0.9	& 61.0	& 1.1   & 82.2 	& 2.2	& 43.9	  & 7.8   & 54.0  & 13.6 &51.5 	& 8.8   & 18.0 	& 11.8 \\
		DEM~\cite{zhang2017learning}&35.0	& 11.1  & 79.4 	& 19.4	&\textbf{67.1} & 30.5  & 86.4 	& 45.1	&51.7	  & 19.6  & 57.9  & 29.2 &40.3 	& 34.3  & 20.5 	& 25.6 \\
		RelationNet~\cite{sung2018learning}&-	& - & - & -	& 64.2	& 30.0  & \textbf{93.4} & \textbf{45.3} & 55.6  & 38.1  & 61.1  & 47.0 &-	& -	& - & - \\		PSR-ZSL~\cite{annadani2018preserving}&38.4	& 13.5 	& 51.4 		& 21.4	&63.8	& 20.7  & 73.8 	& 32.3	& 56.0  & 24.6  & 54.3  & 33.9 &\textbf{61.4}& 20.8 & 37.2 & 26.7 \\
		SP-AEN~\cite{chen2018zero}&- & 13.7  & 63.4 		& 22.6	&-		& 23.3  & 90.9 	& 31.1	&-	  & 34.7  & 70.6  & 46.6 &-	 	& 24.9  & 38.2 	& 30.3 \\
		CAPD~\cite{rahman2018unified}&39.3	& 26.8	&59.5  & 37.0	&52.6		& 	45.2    & 68.6 	& 54.5	&53.8 & \textbf{41.7}  & 44.9 & 43.3 &49.7 & 27.8	& 35.8 	& 31.3 \\
		GDAN~\cite{huang2018generative}	&-	& \textbf{30.4} & 75.0 & \textbf{43.4}	&-	& \textbf{33.2}  & 67.5 	& 44.6	&-	  & 39.3  & 66.7  & \textbf{49.5} &-	& \textbf{38.1}  & \textbf{89.9} & \textbf{53.4} \\
		\midrule
		\textbf{Our TGG}& \textbf{63.5}	& \textbf{58.3} & \textbf{89.6} & \textbf{70.6} & \textbf{77.2} & \textbf{69.8} & 90.1 & \textbf{78.7} & \textbf{64.1} & \textbf{53.8} & \textbf{77.2} & \textbf{63.4} &	\textbf{68.9} & \textbf{65.8} & 88.2 & \textbf{75.4} \\
		
		\bottomrule
	\end{tabular}
	\vspace{-0.6em}
\end{table*}

\subsection{Benchmark datasets}
Following the recently proposed experimental settings~\cite{xian2017zero} for ZSL, we evaluate our TGG on four benchmark datasets: aPY~\cite{Farhadi2009Describing}, AwA2~\cite{xian2017zero}, CUB~\cite{wah2011caltech} and SUN~\cite{sun2012}. Among them, aPY and AWA2 contain coarse-grained classes and are of small and medium size respectively, while both CUB and SUN are medium-size datasets with fine-grained classes. The statistics of them and the associated data splits applied in this paper are provided in Table~\ref{tab:datasets}.

\subsection{Implementation details}

\subsubsection{Image feaures and side information}
For a fair comparison, we use 2048-dim image features from top-layer pooling units of the 101-layered ResNet~\cite{he2016deep} provided by~\cite{sung2018learning}. As for side information, we use continuous valued semantic attributes provided by~\cite{sung2018learning}, whose dimensions are shown in Table~\ref{tab:datasets}. It is noted that our graph generation algorithm is feature-agnostic for both visual feature and side information.

\subsubsection{Network architecture and training settings}
Our aggregate network applies 2 search depth (i.e., 2-hops) with output dimension of 1024 and 512, respectively. We perform batch normalization after each output layer, followed with ReLU activation function. As for multi-head attention module, two dense layers respectively followed by tanh and LeakyReLU~\cite{xu2015empirical} activations are developed for both class-level and instance-level attention. In the relation kernel module, we use a two-layer MLP with batch normalization and ReLU activation for adjacency matrix building, whose input and output dimensions are consistent with the output of the aggregate network and adjacency matrix size, respectively. GCN module is composed of 2 graph convolutional layers with output channel dimensionality of 512 and 128, respectively.
Our whole TGG model is trained end-to-end via ADAM~\cite{kingma2014adam} optimizer with learning rate 0.001 and weight decay 0.0005. The batch size is set to be 128 for all datasets and we use validation sets for early stopping. Both $\lambda_1$ and $\lambda_2$ in Eq.(\ref{eq:total_loss}) are set to be 0.5. We implement our TGG by PyTorch\footnote{https://pytorch.org/} and the source code of our work is available at: \url{https://github.com/zcrwind/tgg-pytorch}.

\subsection{Evaluation metrics}
We follow the standard evaluation metrics used in the literature. For ZSL and FSL, we evaluate the classification performance by the top-1 accuracy, which equals to the percentage of the predicted labels that match the ground truth labels. For GZSL setting, we use Harmonic Mean (HM) of
the separately computed accuracies of seen and unseen classes ($acc_s$ and $acc_u$ respectively), as proposed in~\cite{xian2017zero} as follows:
\begin{displaymath}
HM=\frac{2\times acc_s\times acc_u}{acc_s+acc_u}
\end{displaymath}
The main motivation behind HM is that it can estimate the inherent biasness of GZSL methods towards seen classes. That is, classification methods biased to seen classes will lead to that $acc_s$ is much higher than $acc_u$, and thus the HM value drops down significantly.
For fair comparison, we report the average results of 10 random trails for ZSL, GZSL and FSL.

\subsection{Results and analysis}

\subsubsection{ZSL and GZSL}
We compare our TGG with recent state-of-the-art methods on ZSL and GZSL, and the results are reported in Tabel~\ref{tab:ZSL_GZSL_results}. It is clear that our TGG consistently yields substantial improvements on all datasets for both ZSL and GZSL. More impressively, \textbf{with respect to unseen classes in GZSL setting on several datasets, the accuracy of our TGG is almost 2 times as that of the second place methods}. For example, we respectively achieve the highest $Acc_U$ of 69.8\%/65.8\% on AwA2/SUN for GZSL setting, which has a relative improvement of almost 200\% over the second place GDAN~\cite{huang2018generative}, whose associated accuracies are 33.2\% and 38.1\% respectively. Although our accuracies for seen classes are slightly lower than RelationNet~\cite{sung2018learning} and GDAN~\cite{huang2018generative} on AwA2 and SUN respectively, we still obtain 78.7\% and 75.4\% Harmonic Mean (HM) on these two datasets for GZSL, which are respectively 33.4\% and 24.0\% higher than the two compared methods. This indicates that our TGG can reduce classifier bias to seen classes, and thus manage the trade-off between seen and unseen domains. We attribute this to the introduction of explicit relations and proximity structure modeling, which is vital for alleviating domain shift. Moreover, our TGG surpasses some recent generative methods like SP-AEN~\cite{chen2018zero}, PSR-ZSL~\cite{annadani2018preserving} and GDAN~\cite{huang2018generative} in both unseen class accuracy and HM score, as our TGG explicitly generates relations of graph topology and is robust to noise in the synthesized dummy feature, while retaining the advantages of supervised training. 

\begin{table}
	\caption{FSL results evaluated on four benchmark datasets.}
	\label{tab:FSL_results}
	\begin{tabular}{c|r|cccc}
		\toprule
		k-shot &  Methods & \textbf{aPY} & \textbf{AwA2} & \textbf{CUB} & \textbf{SUN} \\
		\midrule
		\multirow{4}{*}{1-shot} & DeViSE~\cite{frome2013devise} 	&- 		& 81.1 	& 54.9 	& -	\\
		& CMT~\cite{socher2013zero} 		&- 		& 85.6	& 57.3	& - \\
		& CAPD~\cite{rahman2018unified}	 	&71.2 	& 81.4	& 46.3 	&53.7	\\
		& \textbf{Our TGG} 					&\textbf{73.9} 	& \textbf{86.8} & \textbf{65.5}	& \textbf{66.0}\\
		\midrule
		\multirow{4}{*}{3-shot}	& DeViSE~\cite{frome2013devise} 	&- 		& 83.8	& 55.7	& - 		\\
		& CMT~\cite{socher2013zero} 		&- 		& 86.9 	& 58.4	& -		\\
		& CAPD~\cite{rahman2018unified}		&83.6   & 86.9	& 56.9	& 66.3	\\
		& \textbf{Our TGG} 					& \textbf{84.7}	& \textbf{88.1}	& \textbf{69.6}	& \textbf{70.2} \\
		\bottomrule
	\end{tabular}
	\vspace{-1.3em}
\end{table}
\subsubsection{FSL}
Our TGG can be seamlessly extended to FSL, by replacing the dummy data with real support data in unseen classes. In few/one-shot settings, we follow CAPD~\cite{rahman2018unified} to randomly choose three/one instances per unseen class as labeled examples in training. The comparison results are provided in Table~\ref{tab:FSL_results}. Still, our TGG outperforms all the compared methods on all datasets with a quite large margin. Comparing the results of Table~\ref{tab:FSL_results} and Table~\ref{tab:ZSL_GZSL_results}, we can observe that adding real image feature of unseen classes can be always beneficial, and our TGG gains considerable improvement although the given support data is rare. More interestingly, the 3-shot performance in AwA2 dataset (88.1\%) tends to approach that of seen classes in GZSL (90.1\%). This phenomena indicates that our graph generation approach can cope with the data imbalance well, with the aid of the class-level graph and attention-based aggregation.

\begin{table}
	\caption{Ablation studies with ZSL setting on four datasets.}
	\vspace{-0.8em}
	\label{tab:ablation}
	\begin{tabular}{l|cccc}
		\toprule
		Methods 						& \textbf{aPY} 	& \textbf{AwA2}			& \textbf{CUB} 		& \textbf{SUN} \\
		\midrule
		TGG $-$ aggregation			& 35.6			& 43.3			& 31.5			& 29.4 \\
		TGG $-$ attention				& 58.9			& 70.6			& 59.2			& 61.8	\\
		TGG $-$ GCNs					& 57.4			& 70.7 			& 58.5 			& 60.2	\\
		\midrule
		TGG $-$ graph kernel			& 60.3			& 74.6			& 60.4 			& 61.1	\\
		TGG $-$ dual relation prop		& 62.7			& 75.1			& 62.9 			& 63.0	\\
		\midrule
		\textbf{Our TGG} 				&\textbf{63.5} 	& \textbf{77.2} & \textbf{64.1}	& \textbf{68.9}\\
		\bottomrule
	\end{tabular}
	\vspace{-0.9em}
\end{table}
\subsection{Ablation studies}
We conduct ablation studies on ZSL, to further evaluate the effect of different components in our TGG approach, and the results are exhibited in Table~\ref{tab:ablation}. We design ablative experiments from two perspectives, namely \textit{architecture} and \textit{constraints}. In terms of architecture, we independently remove the aggregation network, multi-head attention and GCNs from TGG framework, corresponding to the first three rows of Table~\ref{tab:ablation}. In terms of constraints, we remove the graph kernel ($Loss_K$) or replace the dual relation propagation ($Loss_d$) with standard label propagation algorithm, corresponding to the 4-th and 5-th rows of Table~\ref{tab:ablation} respectively.

From the experimental results, we can obverse that the accuracies drop drastically when the aggregation module is totally removed. This is mainly because the aggregation operation captures proximity structure from class-level prototype graph, it integrates neighborhood information to revise node embeddings and alleviate domain shift, and thus more robust than direct graph generation over the original image features. Similarly, if we simply use mean aggregator without the multi-head attention in aggregation module, the performance will also be impaired significantly (from ~4\% to ~7\% on four datasets). This illustrates that attention is vital in such $\mathcal{G}_C\rightarrow\mathcal{G}_I$ graph translation procedure, as neighborhood information should be finely screened to tackle information dilution and negative knowledge transfer, as well as cope with noise in the dummy instances for ZSL. Moreover, as shown in the third row of Table~\ref{tab:ablation}, GCNs in our relation kernel module also play a crucial role, as they further refine topologies of the generated graph at instance level and increase nonlinearity.

From the constraints perspective, the results of the 4-th and 5-th rows in Table~\ref{tab:ablation} demonstrate the effects of $Loss_K$ and $Loss_d$, respectively. As a regularization, graph kernel constraint ($Loss_K$) encourages the instance-level graph to be consistent with the class-level graph in local structure, which overcomes overfitting and is especially vital for the datasets with fine-grained classes (such as CUB and SUN). Moreover, dual relation propagation ($Loss_d$) is also beneficial for zero-shot generalization, which stably gains around 2\% improvement on four datasets.

\begin{figure}[htb]
	\begin{center}
		\includegraphics[height=4.3cm]{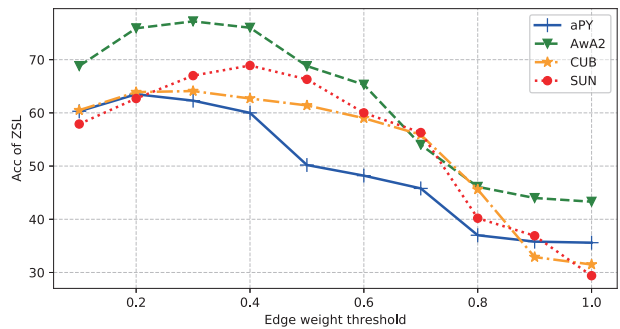}
	\end{center}
	\vspace{-1.0em}
	\caption{Sensitivity experiments of the $\mathcal{G}_C$ size.}
	\label{fig:edge_weight_threshold}
	\vspace{-1.0em}
\end{figure}
\subsection{Sensitivity experiments}
As stated in Section~\ref{sec:class_level_graph_construction}, the initial $\mathcal{G}_C$ is densely connected with normalized edge weights. In order to figure out the effect of $\mathcal{G}_C$ size on the performance of our TGG, we set different thresholds of edge weights to crop $\mathcal{G}_C$, i.e., one edge will be removed if its weight is smaller than the pre-defined threshold. The experiments are conducted on ZSL setting and the results are shown in Figure~\ref{fig:edge_weight_threshold}. We can observe that the size of $\mathcal{G}_C$ is crucial for the subsequent graph generation and classification. Our conclusions are two-folds: (1) Using whole edges can be suboptimal, as some neighbor information with sloppy relations will be involved for graph generation, resulting in negative knowledge transfer. (2) If the edges are removed in large or even in total (when the threshold is set to 1.0), the performance drops drastically. This indicates the fact that prototype relations in $\mathcal{G}_C$ play a vital role for our instance-level graph generation.

%
%
%

\section{Conclusion}
In this paper, we have proposed a unified and flexible framework TGG for ZSL, GZSL and FSL via graph generation, towards a comprehensive relation modeling and utilization in an explicit manner. Our TGG not only accounts for the structural matching between the semantic space and the visual feature space, but also enriches it with instance-level relation modeling, which captures intra-class variance for better decision boundary learning. Extensive experiments performed on widely-used zero-shot and few-shot datasets attest the superiority of our approach. In the future work, we attempt to model relations with more advanced graph generation techniques, as well as reduce the computational complexities of our TGG for larger scale transfer learning situations.

%
\begin{acks}
This work was supported by the National Natural Science Foundation of China under Grant 61876003. It is a research achievement of Key Laboratory of Science, Techonology and Standard in Press Industry (Key Laboratory of Intelligent Press Media Technology).
\end{acks}

\newpage

%
\bibliographystyle{ACM-Reference-Format}
\balance
\bibliography{bibfile}

%
\appendix

\end{document}